\documentclass{article}

\usepackage{arxiv}

\usepackage[utf8]{inputenc} % allow utf-8 input
\usepackage[T1]{fontenc}    % use 8-bit T1 fonts
\usepackage{hyperref}       % hyperlinks
\usepackage{url}            % simple URL typesetting
\usepackage{booktabs}       % professional-quality tables
\usepackage{amsfonts}       % blackboard math symbols
\usepackage{nicefrac}       % compact symbols for 1/2, etc.
\usepackage{microtype}      % microtypography
\usepackage{lipsum}		% Can be removed after putting your text content
\usepackage{graphicx}
\usepackage[numbers]{natbib}
\usepackage{doi}
\usepackage{amsmath,amssymb,amsfonts}
\usepackage{algorithmic}
\usepackage{algorithm}
\usepackage{graphicx}

\title{Latent Space Reinforcement Learning for Multi-Robot Exploration}

\author{ \href{https://orcid.org/0009-0002-6745-7219}{\includegraphics[scale=0.06]{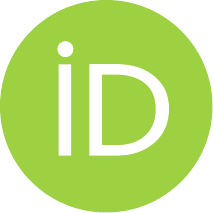}\hspace{1mm}Sriram Rajasekar} \\
	Department of Aerospace Engineering \\
	Indian Institute of Science \\
    Bengaluru, India \\
	\texttt{sriramrajase@iisc.ac.in} \\
	\And
	\href{https://orcid.org/0000-0001-5311-3066}{\includegraphics[scale=0.06]{orcid.pdf}\hspace{1mm}Ashwini Ratnoo} \\
	Department of Aerospace Engineering \\
	Indian Institute of Science \\
    Bengaluru, India \\
	\texttt{ratnoo@iisc.ac.in} \\
}

\renewcommand{\shorttitle}{Latent Space Reinforcement Learning for Multi-Robot Exploration}

\hypersetup{
pdftitle={Latent Space Reinforcement Learning for Multi-Robot Exploration},
pdfsubject={},
pdfauthor={Sriram Rajasekar},
pdfkeywords={},
}

\begin{document}
\maketitle

\begin{abstract}
	Autonomous mapping of unknown environments is a critical challenge, particularly in scenarios where time is limited. Multi-agent systems can enhance efficiency through collaboration, but the scalability of motion-planning algorithms remains a key limitation. Reinforcement learning has been explored as a solution, but existing approaches are constrained by the limited input size required for effective learning, restricting their applicability to discrete environments. This work addresses that limitation by leveraging autoencoders to perform dimensionality reduction, compressing high-fidelity occupancy maps into latent state vectors while preserving essential spatial information. Additionally, we introduce a novel procedural generation algorithm based on Perlin noise, designed to generate topologically complex training environments that simulate asteroid fields, caves and forests. These environments are used for training the autoencoder and the navigation algorithm using a hierarchical deep reinforcement learning framework for decentralized coordination. We introduce a weighted consensus mechanism that modulates reliance on shared data via a tuneable trust parameter, ensuring robustness to accumulation of errors. Experimental results demonstrate that the proposed system scales effectively with number of agents and generalizes well to unfamiliar, structurally distinct environments and is resilient in communication-constrained settings. 
\end{abstract}

% keywords can be removed
\keywords{Autoencoder \and Autonomous mapping \and Decentralized coordination \and Deep Reinforcement learning \and Dimensionality reduction \and Multi-agent systems \and Procedural generation \and Weighted consensus}

\section{Introduction}
\label{sec:introduction}
Autonomous exploration by multiple robots is a fundamental challenge in robotics \cite{quattrini2020}, pivotal for applications where human access is dangerous, impractical, or inefficient. Examples of such applications include search and rescue operations in disaster zones \cite{murphy2014, liu2016}, military reconnaissance, resource prospecting for mining, and planetary exploration for extraterrestrial settlements \cite{wilcox1992}. The primary objective in these scenarios is to achieve rapid coverage of previously unknown spaces, transforming raw sensory data into actionable spatial knowledge in the form of a map \cite{burgard2000}. However, several challenges need to be addressed to reach this goal. Some of the prominent challenges are: safely navigating any given environment including inherently complex and unstructured ones, ensuring the entire system can scale efficiently as the number of robots or the size of the environment increases, and mitigating the effect of errors arising due to noisy communication as well as sensor noise.

Multi-robot exploration strategies designed to tackle some of these challenges have been categorized into centralized and decentralized approaches. Centralized methods, while capable of achieving global coordination by routing all information through a central controller, typically suffer from computational bottlenecks, single points of failure, and heavy reliance on robust, ubiquitous communication \cite{saravanan2020}. Conversely, decentralized methods offer better scalability and robustness to communication failures \cite{talamali2021}. Traditional solutions, whether centralized or decentralized, predominantly rely on hand-crafted heuristics, which are pre-programmed rules or utility functions \cite{burgard2000}, such as directing agents towards the nearest unvisited frontier \cite{yamauchi1997, yamauchi1998}, employing bidding mechanisms based on estimated travel cost versus information gain for task allocation \cite{zlot2006, yan2011}, or utilizing artificial potential fields to enforce spatial separation \cite{reif1999}. Such traditional approaches demand explicit engineering based on domain-specific knowledge such as assumptions about the environmental layout (e.g., corridor widths, room sizes), semantic understanding of the environment \cite{stachniss2008, stachniss2006} and the expected distribution and types of obstacles. This leads to such strategies struggling to generalize when faced with diverse, unstructured environments and often face inherent difficulties in orchestrating complex multi-agent coordination without a global perspective. In contrast to traditional approaches that rely on such hand-crafted heuristics, Deep Reinforcement Learning (DRL) has emerged as a powerful approach \cite{arulkumaran2017}, due to its ability to learn sophisticated, end-to-end policies for complex behaviors directly from high-dimensional sensory inputs and interaction experiences \cite{niroui2019}. The capability of DRL to automatically discover intricate strategies makes it most well-suited for the problem of multi-robot exploration, especially when scalability and generalizability are key requirements.

Despite offering a promising direction, effectively applying a decentralized DRL framework to map-based exploration introduces its own set of significant hurdles, primarily stemming from the nature of the input data. For accurate decision-making and to avoid overlooking critical environmental details such as small navigable gaps or subtle obstacles, high-resolution map inputs are highly desirable, particularly as multi-robot systems are increasingly tasked with exploring large-scale and complex real-world environments where their advantages over single-robot systems are most pronounced \cite{kegeleirs2021}. However, utilizing such high-resolution data directly as input for DRL agents immediately invokes the ``curse of dimensionality,'' which refers to the fact that the vastness of the state space grows exponentially with the input dimensions. This makes it exceedingly difficult for policy networks to learn robust feature extraction concurrently with effective navigational or exploratory behaviors \cite{hu2023}.

Training robust and generalizable DRL agents for large-scale exploration critically depends on exposure to a vast and diverse range of challenging environments. However, the historical difficulty of applying DRL to high-resolution inputs has led many approaches to rely on smaller, less complex map representations. This practical constraint, in turn, has inadvertently resulted in a significant gap in the literature concerning sophisticated and scalable map generation techniques. Since agents were largely confined to simpler inputs, the impetus for developing advanced generators was diminished. Consequently, prior works often utilize simplistic, hand-designed environments or assume pre-existing map sets, which may not adequately capture the variability and complexity of real-world scenarios, as acknowledged in \cite{wang2022}. Some benchmarks, like Procgen \cite{cobbe2020}, employ procedural content generation to create diverse game levels. But to the best of the authors' knowledge, prior works in the field of multi-robot exploration using DRL have not outlined any dedicated and customizable large-scale map generators, offering fine-grained control over features like obstacle density, structure, and navigability. This lack of efficient methods for producing the necessary complex, high-resolution training grounds now presents a considerable bottleneck, significantly hampering the development and rigorous evaluation of DRL solutions capable of tackling realistic, large-scale exploration tasks. Perlin noise \cite{perlin2002} has long been used as a procedural generation technique in fields such as computer graphics due to its ability to produce smooth, natural-looking textures and terrain. It generates continuous gradients of pseudorandom values across a grid, avoiding the discontinuities associated with pure white noise. This is achieved by interpolating between gradients assigned to grid points, resulting in coherent spatial patterns with tunable frequency and scale. Owing to these properties, Perlin noise is a natural candidate for generating training maps that exhibit both randomness and structure.

Further, prior works that attempt to apply RL in larger environments have often resorted to workarounds such as drastically reducing grid resolution, effectively focusing mainly on discrete environments \cite{garaffa2023} which sacrifices crucial fine-grained details. Other works use only small, cropped local map portions as input, which can lead to a loss of essential global context for informed decision-making \cite{bai2017}. This leads to the scalability of prior RL-based frameworks being limited in nature. Consequently, there is a pressing need for methods that can perform effective feature extraction and dimensionality reduction transforming high-resolution inputs into compact, informative representations to make the learning problem tractable for the DRL agents without sacrificing critical information.

In addition, operating effectively in a decentralized multi-agent context demands robust coordination, especially when working in realistically noisy environments. In settings with imperfect sensors or unreliable communication channels, information exchanged between agents can be erroneous, and without mechanisms to critically evaluate and integrate potentially corrupted data, there is a substantial risk of errors propagating throughout the swarm. This leads to degraded collective perception and misguided decision-making, thereby undermining the overall exploration mission. This challenge has foundational roots in probabilistic mapping, where uncertainty is modeled within occupancy grids and information is fused algorithmically \cite{moravec1985}. While these principles address sensor uncertainty for a single agent, they do not directly specify how to optimally fuse entire maps or map segments communicated between multiple agents, which remains an active area of research \cite{weng2020}, especially when the sensed and communicated pieces of information are affected by different types and varying degrees of noise and errors (as will be discussed in Section 5). Thus, there is a need for an information fusion algorithm that allows control over how much trust each individual places on the information received from its neighbors.

Beyond the aforementioned challenges, the inherent complexity and long-horizon nature of autonomous multi-robot exploration present significant difficulties for traditional single-policy DRL systems. Single-policy approaches often struggle to simultaneously master high-level strategic decision-making and fine-grained low-level motion control, leading to sample inefficiency and suboptimal performance in extensive tasks. Complex, long-horizon tasks like multi-robot exploration often benefit from a hierarchical control structure. This motivates exploring hierarchical control structures capable of decomposing the problem into more manageable levels of abstraction.

To address the aforementioned challenges and outline a generalizable, scalable, noise-resilient solution to the problem of multi-robot exploration, this paper introduces a novel decentralized DRL framework incorporating several key innovations. Firstly, we present a sophisticated procedural map generation algorithm, leveraging layered Perlin noise and a controllable difficulty parameter, capable of producing diverse, large-scale and high-resolution environments essential for comprehensive DRL training. Secondly, to overcome the curse of dimensionality associated with such rich map inputs, we pre-train a convolutional autoencoder for effective feature extraction and dimensionality reduction, providing compact state representations to the RL agents. Thirdly, our approach facilitates robust decentralized coordination through a novel map fusion mechanism where agents exchange encoded representations of their independently maintained maps, integrating this information using a configurable trust parameter to mitigate the impact of noisy sensor data and communication as well as errors due to encoding-decoding cycles. Finally, we adopt a hierarchical reinforcement learning architecture \cite{kulkarni2016, tan2023} to manage the complexity of the exploration task, decomposing it into strategic high-level decision-making and low-level navigation. This framework, by design, promotes implicit coordination as agents consider neighbor states in their decision-making at both levels, leading to a scalable and adaptive solution for autonomous exploration in unknown environments.

The remainder of the paper is organized as follows: Section 2 formulates the problem. Details of the proposed map generation algorithm are included in Section 3, while Section 4 presents the training and use of the autoencoder-based feature extraction method, to enable working with large occupancy grids. Section 5 discusses the flow and fusion of information between the agents. The Hierarchical RL framework that is used for training and the training loop itself are described in Section 6. Finally, the experiments and their results are presented in Section 7, and Section 8 concludes the work with remarks on possible future work.

\section{Problem Formulation}

Consider $N$ robots ($A_1, A_2, \dots, A_N$) collaborating to fully explore a given unknown, static environment. They are initiated simultaneously at random positions within the unknown environment in a non-overlapping manner. Each robot $A_i$ has a sensing radius $R_s$ around it, within which it can perfectly sense obstacles as well as communicate with other robots. The environment to be explored is considered to be a square occupancy grid of size $L \times L$ (henceforth referred to as the ``ground map'', $G$) in which ``1'' represents an obstacle and ``0'' represents free space. The following assumptions are made about the environment and the robots:

\begin{enumerate}
    \renewcommand{\labelenumi}{A\arabic{enumi}.} 
    \item The environment is static and two-dimensional.
    \item Robots are modeled as point agents, each occupying a single pixel on the grid.
    \item Robots have range sensors that are free of noise and errors.
    \item Robots know their exact locations within the grid at all times.
\end{enumerate} 

Each robot $A_i$ independently maintains multiple binary occupancy grid maps:
\begin{enumerate}
    \item A ``Self Obstacle Map'' ($O_i^s$), storing obstacle locations sensed purely by $A_i$.
    \item A ``Self Exploration Map'' ($E_i^s$), storing locations explored purely by $A_i$.
    \item A ``Fused Obstacle Map'' ($O_i^f$), storing cumulative obstacle locations obtained through sensing as well as communication.
    \item A ``Fused Exploration Map'' ($E_i^f$), storing cumulative locations explored through sensing as well as communication.
\end{enumerate}

The objective is to train a neural network-based solution using RL to be individually used by each robot as a shared policy to guide the robots to collectively explore the environment. The exploration is considered complete when the collective map of all the robots is completed, that is, 
\begin{equation}
    \bigvee_{i=1}^N O_i^f(x, y) = G.
    \label{eq:completion1}
\end{equation}

A condition equivalent to \eqref{eq:completion1} is to check whether the collective exploration map of the robots is entirely filled, that is, 
\begin{equation}
    \forall x, y \in \{1, \dots, L\},\ \bigvee_{i=1}^N E_i^f(x, y) = 1.
    \label{eq:completion2}
\end{equation}

It is more practical to check for \eqref{eq:completion2} compared to \eqref{eq:completion1} since $G$ is initially unknown to the robots. As $100\%$ exploration is infeasible in unknown environments \cite{kriegel2015}, a soft termination threshold of $80\%$ exploration progress is considered. Fig. 1 shows an example scenario of exploration under progress in the simulations.

\begin{figure}[!h]
\centering
\includegraphics[width=12cm]{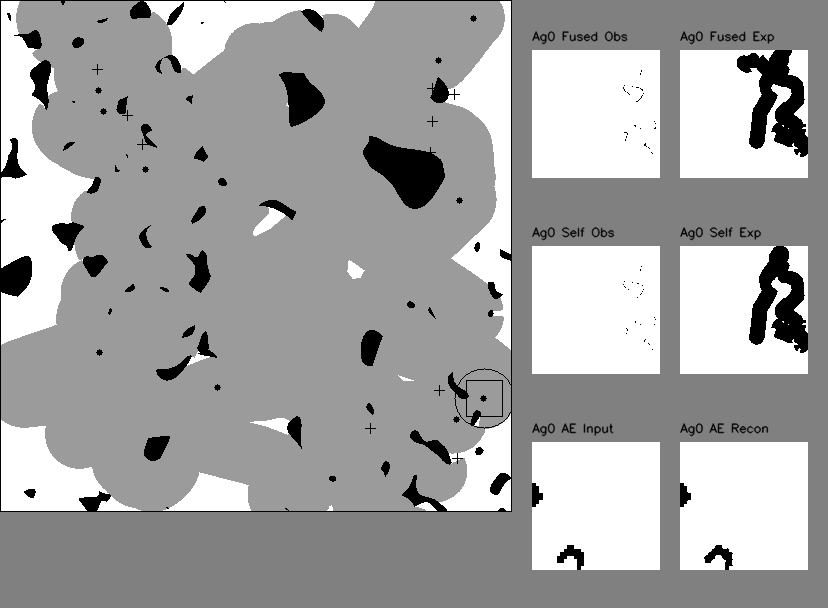}
\caption{An example scenario. (i) To the left: 10 robots exploring the ground map. One agent is highlighted with a circle to visualize the sensing radius. (ii) To the right: From top-left, the highlighted agent's fused obstacle map, fused exploration map, self obstacle map, self exploration map. The final two images are relevant to the autoencoder as discussed in Section 4.}
\label{fig1}
\end{figure}

\section{Perlin Noise-Based Map Generation Algorithm}

The randomly generated environments used in this work are represented as occupancy grids, as previously described. Real-life environments can have unpredictability in terms of shape, scale and size of obstacles. Focusing on unstructured and entirely random environments while generating training data could help in generalizing the solution to any environment, even those unseen in training.

Scaling to larger grids (e.g., $512 \times 512$) with purely random features introduces significant complexity, particularly in generating diverse and meaningful environments suitable for training reinforcement learning (RL) agents. To address this challenge, we propose a fast and flexible map generation algorithm capable of producing randomized environments at scale. The method supports full variability in obstacle size, shape, and density while maintaining computational efficiency. This makes it feasible to generate large training datasets within practical timeframes for use in the learning frameworks described in Sections IV and VI.

Algorithm 1 details the map generation process proposed in this work. We define a single output of Perlin noise as a \textit{layer}, which is generated at a specific spatial frequency, henceforth referred to as the number of \textit{octaves}. Each layer is a $512 \times 512$ grid of real values in the range $[0, 1]$, where values below a chosen threshold (e.g., 0.5) represent free space and values above represent obstacles.  

\begin{algorithm}
\caption{Map Generation Using Masked Perlin Layers}
\label{alg:1}
\begin{algorithmic}[1]
\STATE \textbf{Input:} Map size $L$, crop radius $c$, difficulty $d \in [0, 1]$
\STATE $e \gets 1 - d$ \COMMENT{Ease factor for thresholding}
\FOR{$comp \in \{\text{L}, \text{S}\}$}
    \FOR{$type \in \{\text{base}, \text{mask}\}$}
        \STATE $\textit{oct} \leftarrow \text{Rand}( \textit{comp}{=}\text{L} \ ? \ [7, 12] : [15, 20] )$
        \STATE $\textit{raw} \leftarrow \text{Perlin}(L, \textit{oct})$
        \STATE $layer_{type} \leftarrow \textit{raw} > (type=\text{base} \ ? \ e : 0.5)$
    \ENDFOR
    \STATE $\textit{Map}_{comp} \gets layer_{base} \land layer_{mask}$
\ENDFOR
\STATE $M \leftarrow \text{Perlin}(L, \text{Rand}([1, 3])) > 0.5$
\STATE $\textit{Map} \gets (\textit{Map}_{\text{L}} \land M) \lor (\textit{Map}_{\text{S}} \land \lnot M)$
\STATE Mask center of $Map$ with radius $c$
\STATE Add border to $Map$
\RETURN $\textit{Map}$
\end{algorithmic}
\end{algorithm}

\begin{figure}[!h]
\centering
\includegraphics[width=10cm]{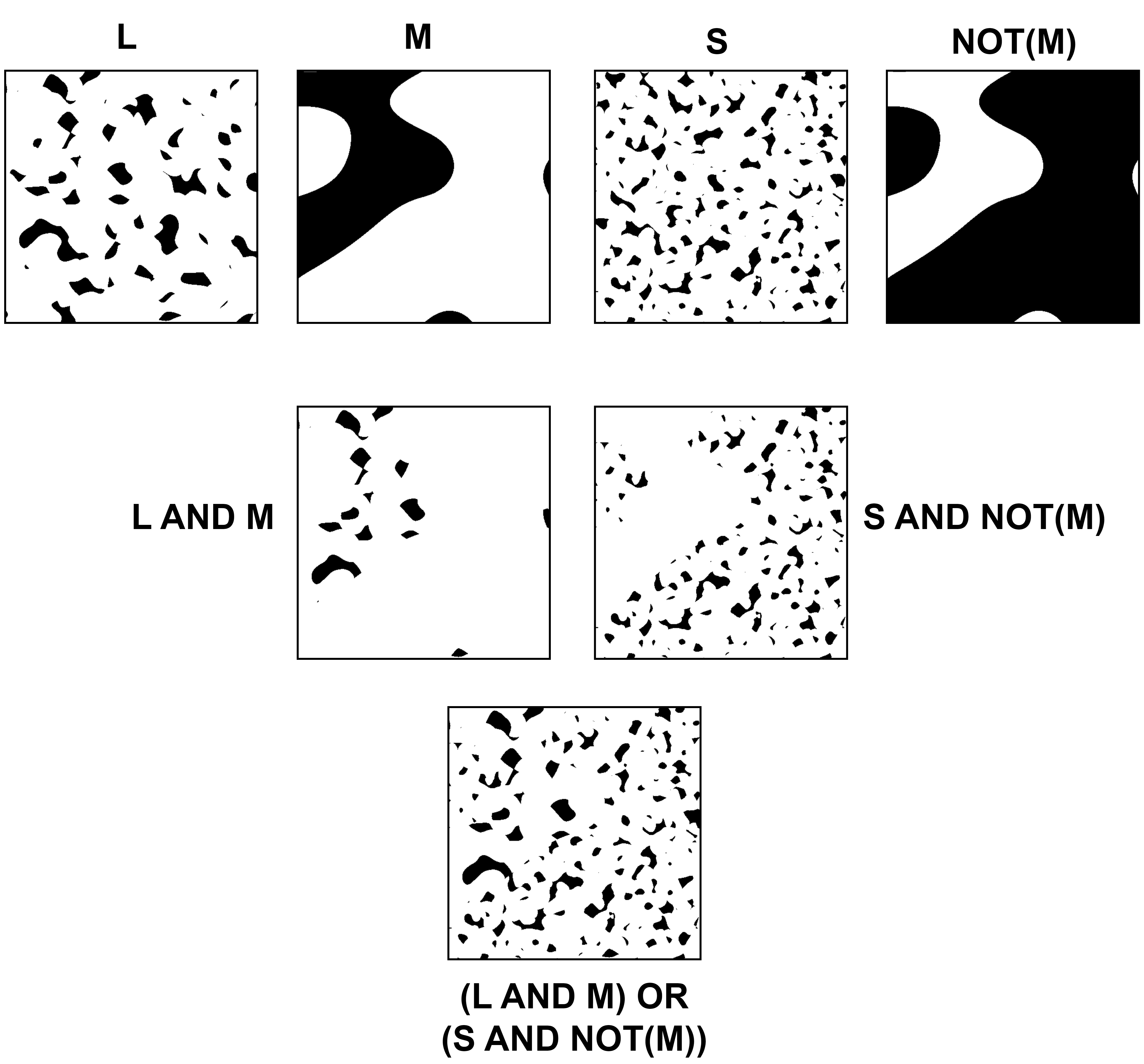}
\caption{An example usage of the map generation algorithm. The labels are as used in Algorithm 1.}
\label{fig2}
\end{figure}

\begin{figure}[!h]
\centering
\includegraphics[width=6cm]{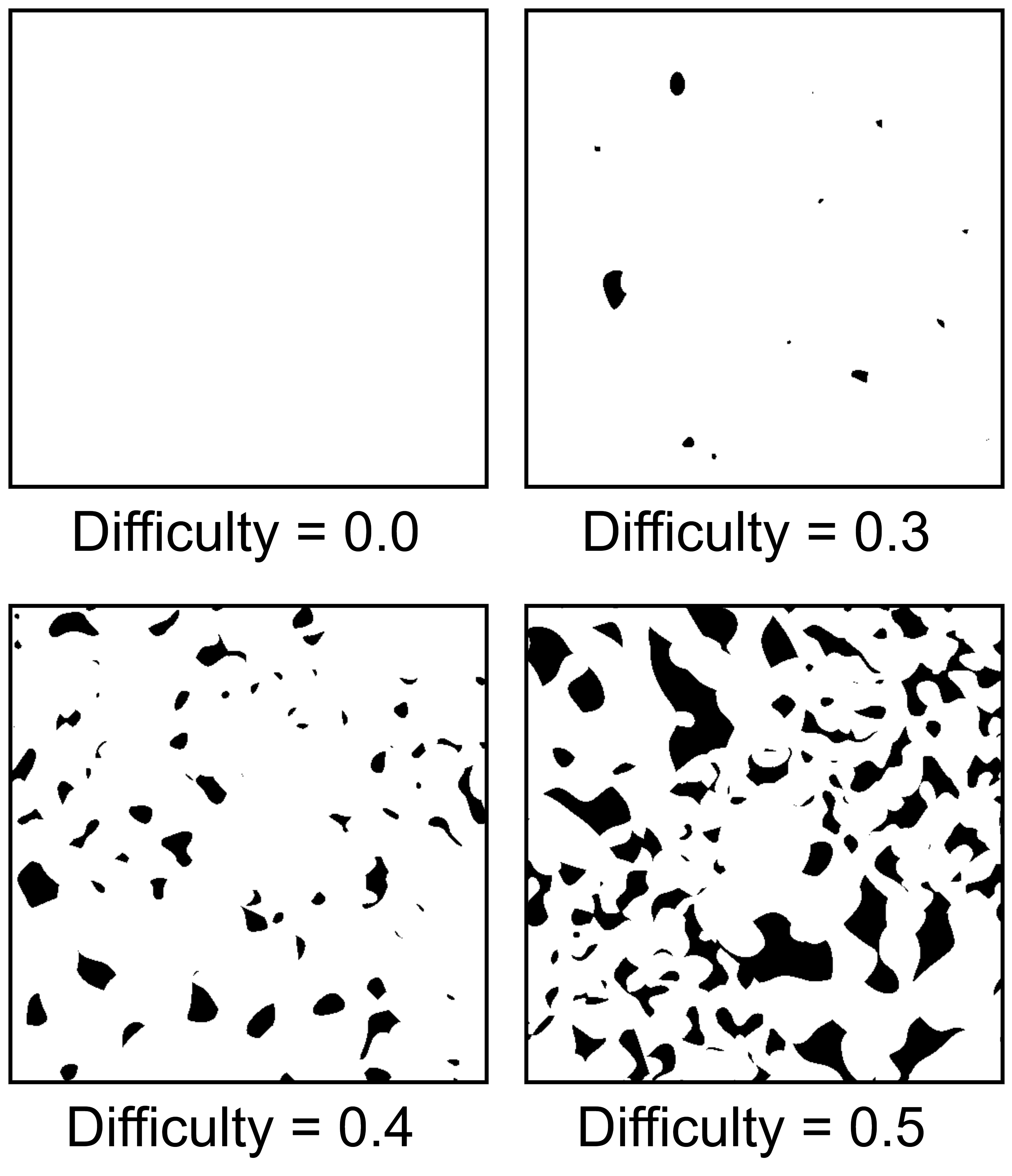}
\caption{Examples of generated maps with varying difficulty levels.}
\label{fig3}
\end{figure}

A single Perlin noise layer, denoted as $layer_{base}$ tends to result in high obstacle density with limited navigable pathways upon thresholding. To alleviate this, a secondary Perlin noise layer, $layer_{mask}$, is generated using a random number of octaves. After thresholding both layers, a binary bitwise AND operation is applied, preserving only those obstacle regions that are present in both $layer_{base}$ and $layer_{mask}$. This effectively introduces more free space into the environment while retaining the organic, structured shapes characteristic of Perlin noise. A controllable parameter $d$ is introduced, which indirectly regulates difficulty by controlling the density of the base layer by thresholding at different levels. A lower value of $d$ leads to a higher threshold, thereby leading to a smaller percentage of the base layer being covered in obstacles. 

While this component map provides improved navigability and obstacle realism, it may lack diversity in obstacle scale. To address this, two such component maps (L and S) are generated using randomly chosen octave values. A third layer M is generated as a pure Perlin noise layer with low octaves to serve as the combination logic for L and S through Algorithm 1. This aggregation results in a map containing obstacles of varying shapes and sizes, with controllable difficulty (density). A central clearing is introduced for initializing robots and virtual borders are added in the form of one-pixel-wide obstacles at the edges. A simplified example of the map generation algorithm is visualized in Fig. 2, and some example maps generated with different difficulty settings are shown in Fig. 3.

\section{Feature Extraction}

Utilizing high-resolution occupancy grids is essential for detecting fine-grained obstacles, yet directly feeding such high-dimensional data to RL agents triggers the ``curse of dimensionality,'' making the simultaneous learning of feature extraction and control policies intractable. Encoders as part of end-to-end trained systems have been used to circumvent this issue, even in relatively smaller scale problems \cite{liu2023}. In an inherently deeper network meant for larger scale problems, end-to-end training might introduce further difficulties. To circumvent this, we employ a convolutional autoencoder to compress high-fidelity maps into compact latent representations, effectively decoupling visual processing from decision-making. This approach builds upon prior works where autoencoders have successfully been used to distill complex inputs into low-dimensional states for robotic control \cite{finn2016}. A visualization of the usage of the autoencoder to compress maps is shown in Fig. 4.

\begin{figure}[!h]
\centering
\includegraphics[width=7cm]{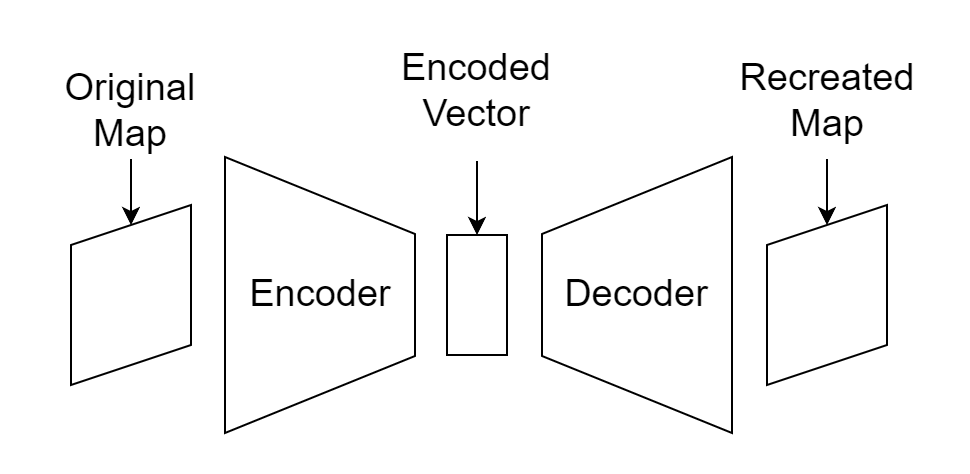}
\caption{A block diagram of a typical autoencoder.}
\label{fig4}
\end{figure}

\begin{table}[!h]
\centering
\caption{Autoencoder Architecture Specifications}
\label{tab:ae_arch}
\begin{tabular}{lc}
\hline
\textbf{Layer} & \textbf{Output Dimensions} \\
\hline
\multicolumn{2}{c}{\textbf{Encoder}} \\
Input & $1 \times 512 \times 512$ \\
Conv2D Block 1 & $16 \times 256 \times 256$ \\
Conv2D Block 2 & $32 \times 128 \times 128$ \\
Conv2D Block 3 & $64 \times 64 \times 64$ \\
Conv2D Block 4 & $128 \times 32 \times 32$ \\
Conv2D Block 5 & $128 \times 16 \times 16$ \\
Conv2D Block 6 & $128 \times 8 \times 8$ \\
Conv2D Block 7 & $128 \times 4 \times 4$ \\
Flatten + Tanh & \textbf{2048 (Latent Vector)} \\
\hline
\multicolumn{2}{c}{\textbf{Decoder}} \\
Unflatten & $128 \times 4 \times 4$ \\
ConvTranspose Blocks 1--6 & \textit{Upsample (Symmetric)} \\
ConvTranspose Block 7 & $1 \times 512 \times 512$ \\
Final Output (Sigmoid) & $\mathbf{1 \times 512 \times 512}$ \\
\hline
\end{tabular}
\end{table}

The autoencoder is trained and tested on 2000 images generated with random difficulty settings using the map generation algorithm outlined in Section 3, with a train-test ratio of 9:1. The autoencoder architecture consists of a symmetric encoder-decoder network designed to process $512 \times 512$ grayscale occupancy grids. The encoder utilizes seven convolutional blocks to progressively compress the spatial dimensions from 
$512 \times 512$ down to $4 \times 4$
, while increasing channel depth from $1$ to $128$. This results in a flattened latent vector of size $2048$, which serves as part of the compressed state input for the RL agents. The detailed architecture specifications are provided in Table~\ref{tab:ae_arch}.

\begin{figure}[!h]
\centering
\includegraphics[width=12cm]{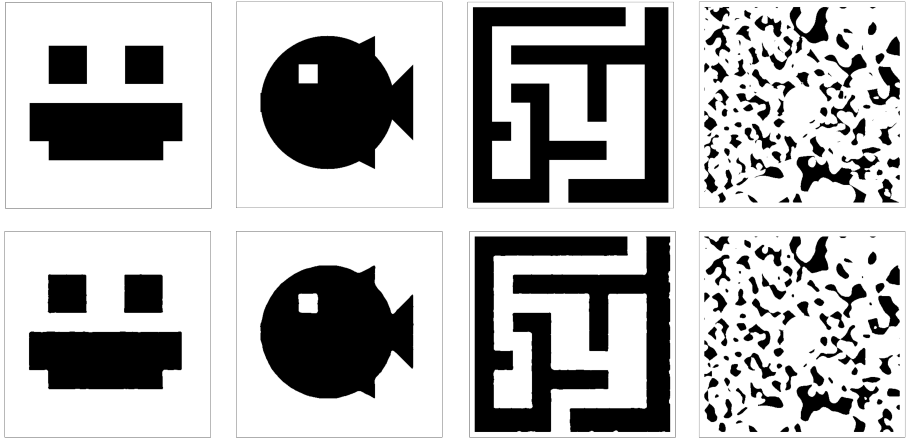}
\caption{Generalization test performed on the autoencoder. The top row displays the original grayscale images fed into the encoder. The bottom row displays the reconstructed images after the encoded latent vectors are fed into the decoder.}
\label{fig5}
\end{figure}

The trained autoencoder is tested for generalization by inputting images that have no visual relation to the training data. The results in Fig. 5 show that the autoencoder has generalized well to encode any black and white image into an informative latent space and then reconstruct it into a recognizable reconstruction. Only the fourth column in Fig. 5 is representative of the training data used for the autoencoder.

\section{Information Flow}

The $m$ robots within the communication radius of a given robot $A_i$ are called its neighbors. Of these neighbors, the robots that $A_i$ communicates with are called communicating neighbors. The maximum number of communicating neighbors is a training parameter and a design choice, labeled $M$.

\begin{algorithm}
\caption{Choosing communicating neighbors $C_1, \dots, C_M$ for agent $A_i$}
\label{alg:2}
\begin{algorithmic}[1]
\IF{$m = 0$}
    \STATE Consider $M$ virtual agents at the opposite quadrant's corner to be the communicating neighbors
\ELSIF{$m < M$}
    \STATE Consider the $m$ nearest agents as communicating neighbors along with $M-m$ virtual agents at the opposite quadrant's corner
\ELSIF{$m \geq M$}
    \STATE Consider the $M$ nearest agents as communicating agents
\ENDIF
\end{algorithmic}
\end{algorithm}

Choosing the $M$ communicating neighbors $C_1, \dots, C_M$ is done based on the relationship between $m$ and $M$ as in Algorithm 2, which ensures that the chosen number of communicating neighbors is always $M$. This is crucial since the information from the communicating neighbors will be part of the inputs to the neural networks, thus requiring the number of communicating neighbors to be consistent throughout. When a certain amount of time has passed since $A_i$ last communicated, it will communicate with the next robot(s) it finds. During a communication event, the robot initiating communication will request and receive the following information from its chosen neighbors: the neighbor's ID ($k$), current position, current velocity, current target, and the encoded vectors of the neighbor's self obstacle map $O_k^s$ and self exploration map $E_k^s$.

While each robot's Self Maps ($O^s, E^s$) only accumulate errors through the robot's individual sensor noise, the Fused Maps ($O^f, E^f$) accumulate errors through not only that robot's sensor noise, but also the sensor noise, communication noise and encoding-decoding errors from each robot it has communicated with, and by extension from each robot that those robots communicated with and so on. To prevent the cascading of errors through the swarm, agents share only their Self Maps rather than their Fused Maps during a communication event. 

The fusion process must integrate these pieces of information (Self and Fused Maps of the robot and the Self Maps obtained from other robots during communication) while mitigating the effects of channel and sensor uncertainty. To achieve this, we employ a weighted consensus mechanism governed by a trust parameter $\beta \in [0,1]$, which modulates the reliance on self-perception versus neighbor data. This approach leverages the redundancy of the swarm to filter out random errors, improving map resilience as the number of agents increases. The fusion algorithm operates on the set of maps available to robot $A_i$ after a communication event: its own current maps ($O_i^s$, $E_i^s$, $O_i^f$, $E_i^f$) and the decoded maps received from neighbors $k \in C_1, \dots, C_M$ ($O_k^s$, $E_k^s$). The following subsections detail how these inputs are combined to compute the intermediate fused maps and then the final fused maps $O_i^s$, $E_i^s$, $O_i^f$, $E_i^f$.

\subsection{Intermediate Map Computation}

Temporary intermediate exploration map ($\hat{E}$) and obstacle map ($\hat{O}$) are first computed based on the current self-perceived and received neighbor maps. Since localization is assumed to be ideal, the self exploration maps offer dependable information. Thus, the intermediate exploration status at coordinate $(x,y)$ is simply the result of the logical OR operation performed over all available exploration maps at those coordinates.
\begin{equation}
    \hat{E}(x,y) = E_i^s(x,y) \lor \bigvee_{k \in C} E_k^s(x, y).
\end{equation}
$\hat{E}$ indicates whether a pixel was considered explored by the self map of the agent or any of its communicating neighbors in the current step.

The computation of the intermediate obstacle value depends on the exploration status of the pixel by the agent and its neighbors. Considering ${C}$ to be the set of IDs of non-virtual communicating neighbors, let $n(x,y) = \sum_{k \in C} E_k^s(x,y)$. Four cases are considered, based on the values of $E_i^s(x, y)$ and $n(x,y)$:

\subsubsection*{Case 1: $E_i^s(x,y)=0$ and $n(x,y) = 0$}
The location $(x, y)$ has not been explored by any of the agents involved in the communication event, including $A_i$. No new information is available regarding the obstacle status of $(x,y)$ from the current event.
\begin{equation}
    \hat{O}(x,y) = 0.
    \label{eq:case4_obs}
\end{equation}

\subsubsection*{Case 2: $E_i^s(x,y)=1$ and $n(x,y) = 0$}
$A_i$ has explored $(x,y)$ while none of the other communicating neighbors in this timestep have. The robot thus relies entirely on its self-perceived obstacle status.
\begin{equation}
    \hat{O}(x,y) = O_i^s(x,y).
\end{equation}

\subsubsection*{Case 3: $E_i^s(x,y)=0$ and $n(x,y) > 0$}
Despite $A_i$ itself not having explored $(x,y)$, some of the communicating neighbors have done so. The robot thus adopts the average obstacle belief from the neighbors that explored this pixel.
\begin{equation}
    \hat{O}(x,y) = \frac{1}{n(x,y)} \sum_{k \in C} O_k^s(x,y).
\end{equation}

\subsubsection*{Case 4: $E_i^s(x,y)=1$ and $n(x,y) > 0$}
$(x,y)$ has been explored by $A_i$ as well as some of the communicating neighbors. The trust weight ratio $W(x,y)$ defined as the agent's self-perception weight relative to the neighbor information weight is calculated based on the number of exploring neighbors $n(x,y)$ and the trust parameter $\beta$ as $W(x,y) = (n(x,y) - 1) \beta + 1$. The effect of the trust parameter is visualized in Fig. 6. It is noted that when $\beta = 0$, $W(x,y) = 1$ implying $A_i$ trusts the data from its communicating neighbors as much as it trusts its own collected data. This represents the maximum trust it can display. On the contrary, when $\beta = 1$, $W(x,y) = n(x,y)$ implying $A_i$'s belief can be overridden only when all the other communicating neighbors agree that $A_i$ is wrong. This represents the minimum trust that $A_i$ can display.

With the calculated trust weight ratio, the intermediate obstacle value is then computed as
\begin{equation}
\begin{split}
    \hat{O}(x,y) = \frac{W(x,y)}{W(x,y) + n(x,y)} O_i^s(x,y) \\ + \frac{1}{W(x,y) + n(x,y)} \sum_{k \in C} O_k^s(x,y).
\end{split}
\end{equation}

\begin{figure}[!h]
\centering
\includegraphics[width=10cm]{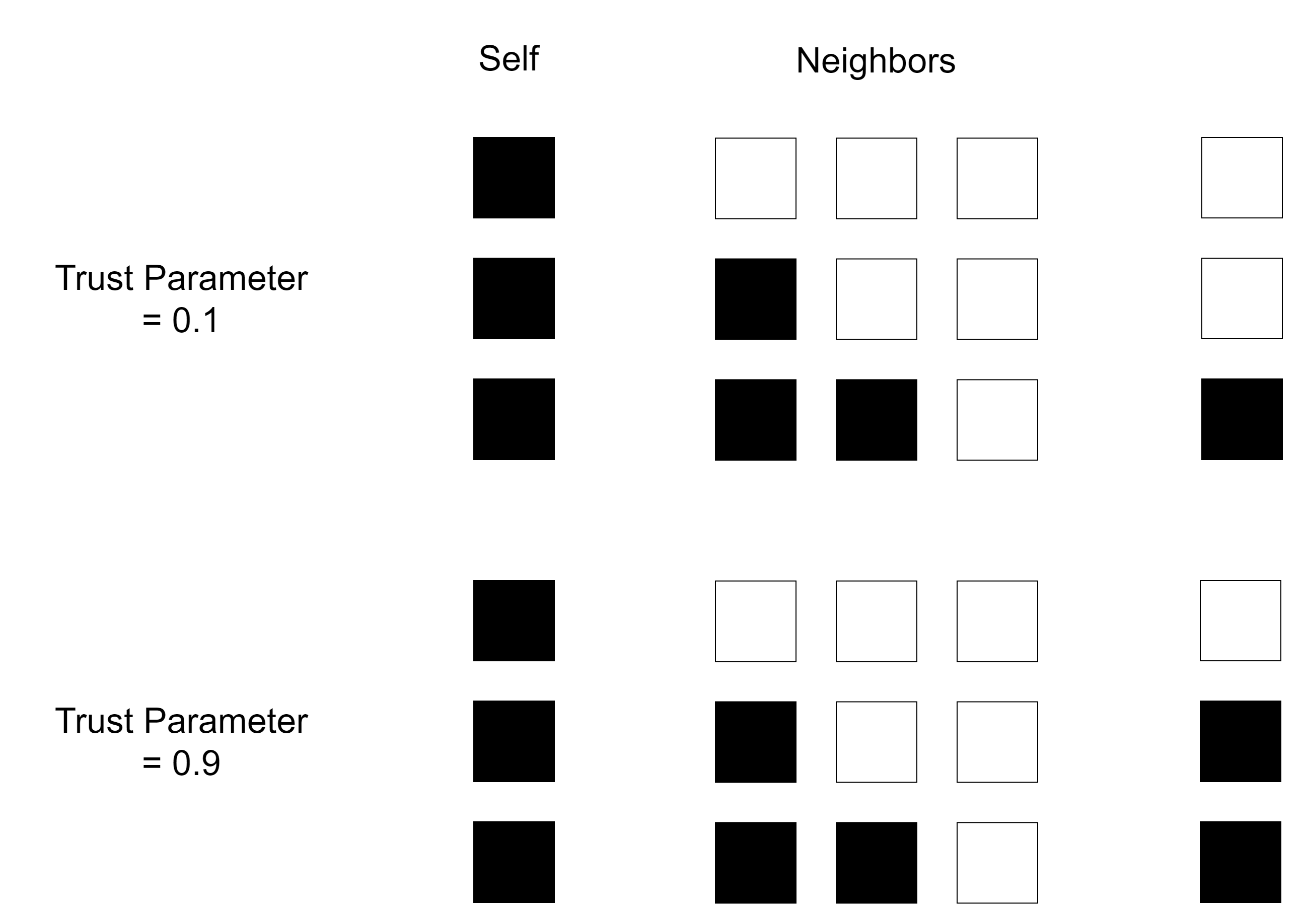}
\caption{An example showcasing the effect of different trust parameters.}
\label{fig6}
\end{figure}

The intermediate floating-point maps $\hat{E}$ and $\hat{O}$ are thresholded at 0.5 to convert them to binary maps. These represent the fused consensus for the current timestep, which does not include any information collected from other neighbors in the past.

\subsection{Thresholding and Temporal Fusion}

The persistent fused maps ($O_i^f$ and $E_i^f$) are updated by integrating them with the new information in $\hat{E}$ and $\hat{O}$.

\begin{itemize}
    \item \textbf{Exploration Map Update:} The updated exploration map simply includes any newly explored areas.
    \begin{equation}
        E_i^f(x,y) = E_i^f(x,y) \lor \hat{E}(x,y).
    \end{equation}
    \item \textbf{Obstacle Map Update:} The obstacle map update prioritizes information from the current fusion step ($\hat{O}(x,y)$) for pixels that were explored in this step ($(x,y)$ such that $\hat{E}(x,y)=1$), while retaining the previous obstacle state ($O_i^f(x,y)$) for pixels not involved in the current fusion update ($(x,y)$ such that $\hat{E}(x,y)=0$).
    \begin{equation}
    \begin{split}
        O_i^f(x,y) = (\hat{O}(x,y) \land \hat{E}(x,y)) \\ \lor (O_i^f(x,y) \land \neg \hat{E}(x,y)).
    \end{split}
    \end{equation}
\end{itemize}

This temporal fusion step ensures that the $A_i$'s fused maps reflect the latest available information while preserving knowledge about areas not covered in the current sensing or communication event. The trust parameter $\beta$ allows tuning the system's sensitivity to potentially noisy neighbor information versus reliance on self-perception.

\section{Reinforcement Learning Framework and Training}

The proposed multi-robot exploration system employs a hierarchical reinforcement learning approach, where the overall exploration task is decomposed into two sub-problems: (1) selecting strategic exploration waypoints, and (2) navigating to these selected waypoints efficiently and safely. To this end, two distinct neural network policies, referred to as Network 1 (Exploration Policy) and Network 2 (Navigation Policy) are trained. This section details the training environment, the navigational policy network, the exploration policy network, and the reward structure designed to elicit effective behavior from the networks.

\subsection{Multi-Agent Reinforcement Learning Environment}

A custom multi-agent environment was developed to simulate the robot exploration task. The environment simulates $N$ robots operating within a $512 \times 512$ grid-based map generated by the algorithm described in Section 3. Key environment and robot parameters used during training are summarized in Table~\ref{tab:env_params}.

\begin{table}[!h]
    \caption{Key Parameters}
    \label{tab:env_params}
    \centering
    \begin{tabular}{ll}
        \hline
        Parameter & Value \\
        \hline
        Number of robots ($N$) & 10 \\
        Max Communicating Neighbors ($M$) & 3 \\
        Map Size & $512 \times 512$ pixels \\
        Sensing Radius ($R_s$) & 30 pixels \\
        Robot Collision Radius ($col\_rad$) & 3 pixels \\
        Maximum Robot Velocity ($max\_vel$) & 10 pixels/s \\
        Maximum Robot Acceleration ($max\_acc$) & 5 pixels/s$^2$ \\
        Map Difficulty ($d$) & 0.1, 0.2, 0.3, 0.4 \\
        Trust Parameter for Map Fusion ($\beta$) & 0.8 \\ 
        Communication Frequency ($comm\_freq$) & 1 time step\\
        Time Step ($dt$) & 0.1 s \\
        \hline
    \end{tabular}
\end{table}

Each robot possesses a simulated range-sensor for obstacle detection and updates its individual maps ($O_i^s, E_i^s$). Communication between robots allows for the exchange of encoded self-maps, which are then fused into each robot's fused maps ($O_i^f, E_i^f$) using the trust-based mechanism detailed in Section 5. The autoencoder described in Section 4 is utilized for encoding maps during inter-agent communication.

\subsection{Navigation Policy Network}

The navigation policy network is responsible for safely guiding the robot around obstacles and for implicitly learning behavior that prevents inter-agent collisions while efficiently reaching a chosen target waypoint. The action space is continuous, representing the desired acceleration vector $(\ddot{x}_i, \ddot{y}_i)$. Each component of the acceleration is clipped in such a way that the magnitude is within the set maximum acceleration while maintaining the direction selected by the network. The observation for each agent $A_i$ for its navigation policy network is a continuous vector comprising the following components:
\begin{enumerate}
    \item A 2048-dimensional latent vector generated by passing a cropped and processed version of $A_i$'s fused obstacle map ($O_i^f$) through the pre-trained encoder (Section 4). The crop is centered on the agent and has a radius of $0.6 \times R_s$. The obstacles are dilated to help the encoder during early stages when information is sparse on the obstacle map.
    \item A 4-dimensional vector representing $A_i$'s current position $(x_i, y_i)$ and velocity $(v_{x,i}, v_{y,i})$, normalized by map size and maximum velocity respectively.
    \item A 2-dimensional vector representing the current navigation target $(t_x, t_y)$ for $A_i$, normalized by map size. For training the navigation policy network before the exploration policy network, this target is randomly assigned by the environment.
    \item For each of the $M$ communicating neighbors (real or virtual, as per Algorithm 2), a 4-dimensional vector representing their normalized position and velocity. This results in a $4 \times M$ dimensional vector.
\end{enumerate}

The components are concatenated to form a single observation vector, with values of some elements normalized to the range $[-1, 1]$ and others to $[0, 1]$. The Navigation Policy utilizes a Multi-Layer Perceptron (MLP) architecture for both its actor and critic components. The policy network (actor) and the Q-function networks (critics) each consist of four hidden layers with [1024, 512, 512, 128] units respectively, employing LeakyReLU activation functions. The reward function for the navigational policy is designed to encourage reaching its target efficiently while penalizing collisions and risky behaviors. For each agent $A_i$ at each timestep, the reward $R_i$ is a sum of several components:

\begin{enumerate}
    \item A large positive reward ($+1500$) is given if the agent reaches its current target (defined as being within a 15-unit radius).
    \item A reward proportional to the reduction in distance to the target compared to the previous step ($4 \times \Delta d_{target}$). This encourages continuous progress.
    \item A reward based on the cosine similarity between the agent's velocity vector and the vector towards the target ($2 \times (\cos \theta)^3$). This encourages heading towards the target, along with the distance reduction term.
    \item A negative reward ($-100$) if the agent collides with a static obstacle (if its current velocity leads to being within the collision radius to an obstacle in next step) or another robot (if its current velocity leads to being within less than twice the collision radius distance of another robot).
    \item A negative reward based on predicted future collisions within a short time horizon ($T_{coll\_window}=15$ steps). This penalty is scaled by how soon the collision is predicted to happen:
    \begin{equation}
        R_{predcoll} =  S_{coll} \times \frac{T_{coll\_window} - t_{coll} + 1}{T_{coll\_window}},
    \end{equation}
    where $t_{coll}$ is the predicted time to collision with an agent or obstacle ($t_{coll} \leq T_{coll\_window}$), and $S_{coll}$ is a negative scaling factor ($-10$).
\end{enumerate}
The total reward for agent $A_i$ is $R_i = R_{term} + R_{prox} + R_{align} + R_{coll} + R_{predcoll}$.

\subsection{Exploration Policy Network}

The exploration policy network is responsible for effective exploration through waypoint selection and for implicitly learning behavior that enables intermittent contact with neighbors to communicate. The action space is continuous, representing the desired waypoint as $x$ and $y$ coordinates normalized by the map dimensions. The observation for each agent $A_i$ for its exploration policy network is a continuous vector whose components are:

\begin{enumerate}
    \item A 2048-dimensional latent vector generated by passing $A_i$'s fused exploration map ($E_i^f$) through the pre-trained encoder (Section 4).
    \item A 4-dimensional vector representing $A_i$'s current position $(x_i, y_i)$ and velocity $(v_{x,i}, v_{y,i})$, normalized by map size and maximum velocity respectively.
    \item For each of the $M$ communicating neighbors (real or virtual, as per Algorithm 2), a 4-dimensional vector representing their normalized position and velocity. This results in a $4 \times M$ dimensional vector.
\end{enumerate}

The components are concatenated to form a single observation vector, with values of some elements normalized to the range $[-1, 1]$ and others to $[0, 1]$. The Exploration Policy utilizes a Multi-Layer Perceptron (MLP) architecture for both its actor and critic components. The policy network (actor) and the Q-function networks (critics) each consist of three hidden layers with [1024, 256, 64] units respectively, employing LeakyReLU activation functions. The reward function for the exploration policy is designed to encourage exploring efficiently while penalizing wasted time and picking inefficient waypoints. For each agent $A_i$ at each timestep, the reward $R_i$ is a sum of several components:

\begin{enumerate}
    \item A large positive reward at the end of the episode, scaled by global fractional exploration progress ($f_{exp}$) as $100 \times (2 \times f_{exp} - 1)$ to encourage a global aspect to exploration.
    \item A reward proportional to the increase in fractional self-explored progress ($f_{self\_exp}$) in this step as compared to the previous step, scaled as $(200 + 300 \times f_{exp}) \times f_{self\_exp}$. This is to encourage continuous progress with special emphasis on last-mile exploration.
    \item A negative reward ($-3$) if the agent picks the next waypoint to be very near to ($\leq 40 $ pixels from) the current location, and a positive reward ($+1.5$) otherwise.
\end{enumerate}
The total reward for agent $A_i$ is the sum of these individual components.

\subsection{Hierarchical Training Procedure}
The overall training follows a two-stage hierarchical process:
\begin{enumerate}
    \item During the navigation policy's training, valid random targets are randomly assigned to each robot by the environment. The primary objective here is to learn robust collision avoidance and target-reaching behaviors.
    \item Then the weights of the navigation network are frozen, and the exploration policy is trained. The exploration network sees every 50 steps of the navigation network as one macro step. The primary objective is to pick a target in each macro step to maximize exploration.
\end{enumerate}

\begin{figure}[!h]
\centering
\includegraphics[width=5.9cm]{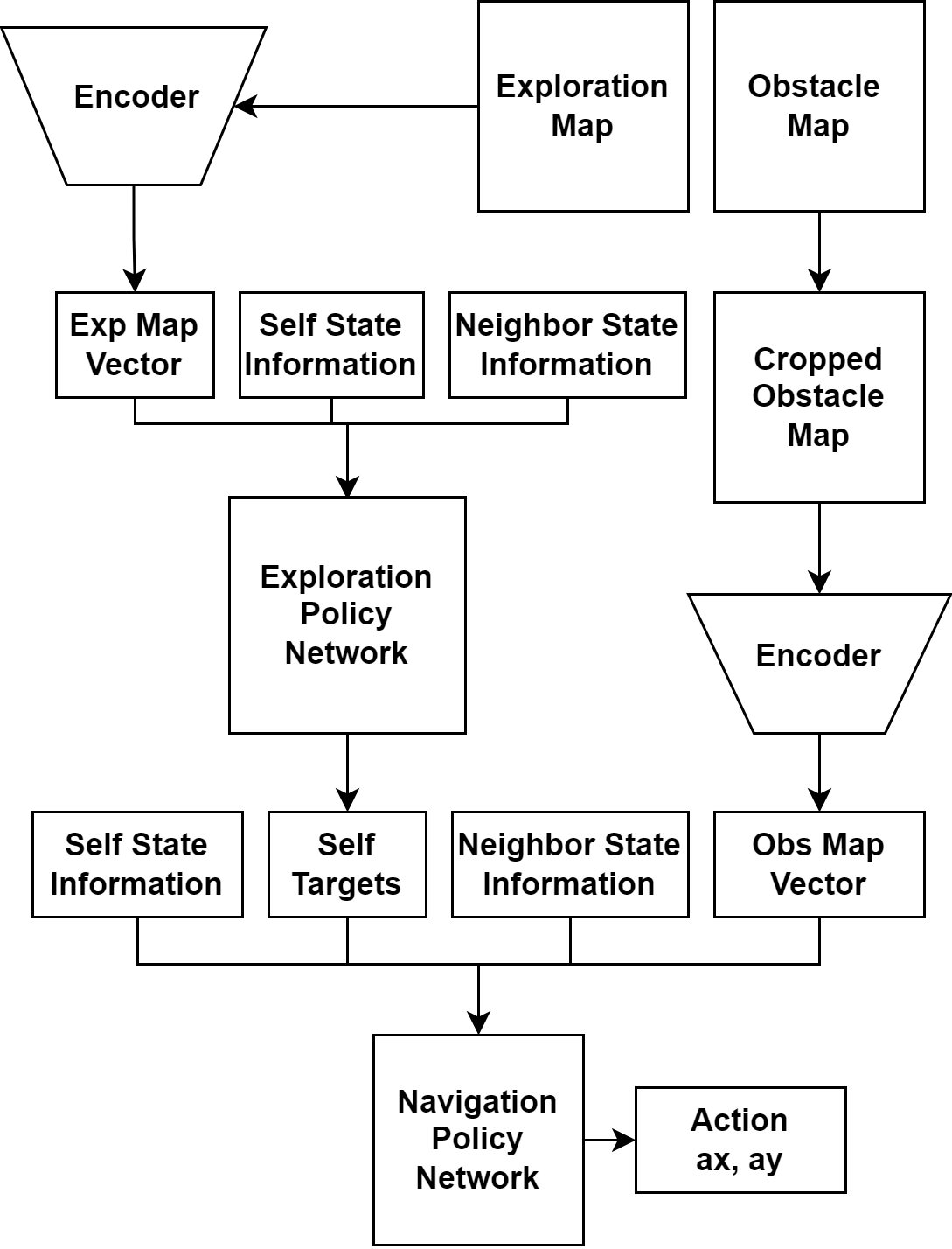}
\caption{An overview of the proposed hierarchical framework.}
\label{fig7}
\end{figure}

This staged approach simplifies the learning problem by allowing each network to specialize in its respective task. Both networks are trained using the Soft Actor-Critic (SAC) algorithm \cite{haarnoja2018}. While prior hierarchical exploration frameworks in continuous scenarios have predominantly utilized Deep Deterministic Policy Gradients (DDPG) \cite{garaffa2023}, we select SAC for its maximum entropy formulation. Unlike DDPG, which is deterministic and prone to converging on suboptimal local minima without heuristic noise injection, SAC's stochastic policy naturally encourages diverse exploratory behaviors. Furthermore, this inherent stochasticity offers greater resilience to the non-stationarity typical of decentralized multi-agent environments compared to brittle deterministic policies. The training in both cases is conducted through a pooled experience replay buffer that treats each episode with N agents as N episodes from the perspective of a single agent. This is for effective generalization of a shared policy that can be cloned onto any number of agents moving forward. The fully trained system, with both networks functioning independently within each agent, is visualized in Fig. 7.

\section{Results}
\label{sec:results}

We evaluate the trained policy networks on scalability, generalization, and noise resilience. Performance is assessed based on exploration completion time for familiar and unfamiliar maps, and final map accuracy under noisy communication conditions.

\subsection{Performance on Familiar Maps}

\begin{figure}[!h]
\centering
\includegraphics[width=8cm]{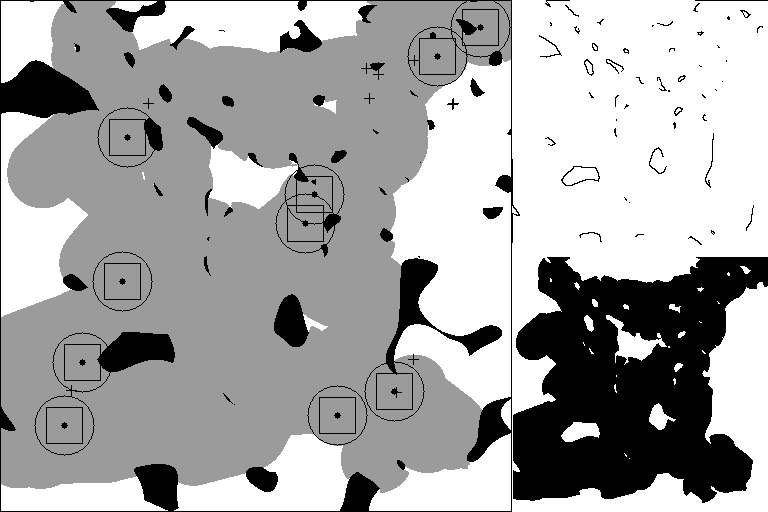}
\caption{Exploration on a familiar map. Left: Ground map with gray explored regions. Top Right: Cumulative obstacle map. Bottom Right: Cumulative exploration map.}
\label{fig8}
\end{figure}

\begin{figure}[!h]
\centering
\includegraphics[width=12cm]{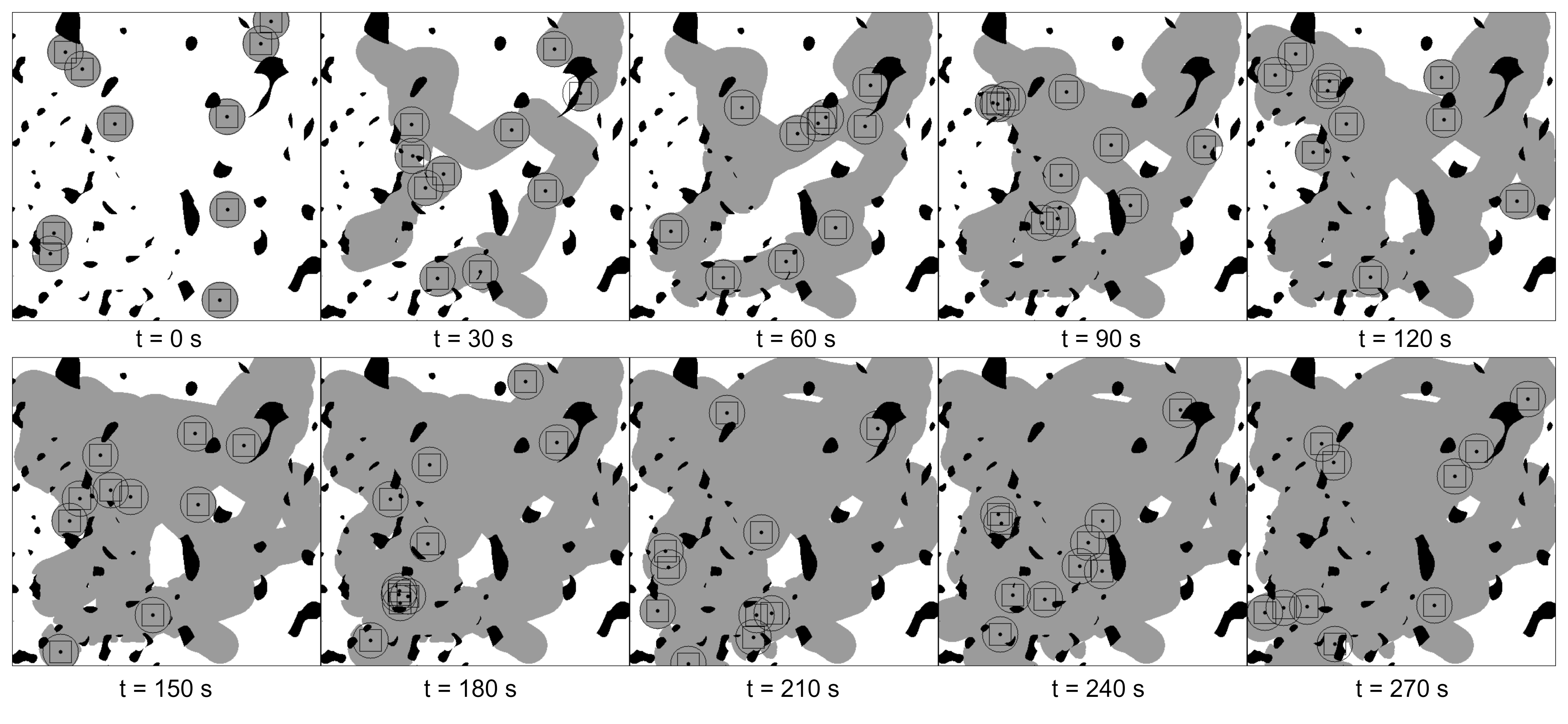}
\caption{Snapshots of exploration progress. $N = 10$, $d = 0.4$.}
\label{fig9}
\end{figure}

Familiar maps are generated using the algorithm from Section 3, matching the training data distribution. Fig. 8 is a visualization of a typical simulation episode while exploration is underway. We conducted 10 simulation episodes for each combination of agent counts and map difficulties listed in Table~\ref{tab:sim_params}. Episodes conclude upon reaching 80\% exploration or a 9000-step (15 min) limit. Fig.~\ref{fig9} illustrates the time evolution of a sample run.

\begin{table}[!h]
\centering
\caption{Evaluation Parameter Settings}
\label{tab:sim_params}
\begin{tabular}{ll}
\hline
\textbf{Parameter}       & \textbf{Value}             \\
\hline
Number of Robots         & 1, 5, 10, 15, 20  \\
Difficulty Level         & 0.1, 0.2, 0.3, 0.4 \\
Max Communicating Neighbors & 3 \\
Map Size & $512 \times 512$ pixels \\
Sensing Radius & 30 pixels \\
Robot Collision Radius & 3 pixels \\
Maximum Robot Velocity & 10 pixels/s \\
Maximum Robot Acceleration & 5 pixels/s$^2$ \\
Trust Parameter ($\beta$) & 0.8 \\ 
Time Step & 0.1 s \\
Communication Frequency & 1 time step\\
Maximum Time Allowed & 9000 time steps (15 min)\\
Episodes per Parameter Set & 10\\
\hline
\end{tabular}
\end{table}

\begin{table}[!h]
\centering
\caption{Average Time (s) for simulations in familiar maps}
\label{tab:avg_time}
\setlength{\tabcolsep}{4pt}
\begin{tabular}{|c|c|c|c|c|}
\hline
\textbf{N / d} & \textbf{0.10} & \textbf{0.20} & \textbf{0.30} & \textbf{0.40} \\
\hline
1  & 900.00 & 900.00 & 900.00 & 900.00 \\
5  & 305.70 & 321.50 & 322.10 & 391.17 \\
10 & 166.22 & 162.33 & 169.50 & 250.33 \\
15 & 122.00 & 126.00 & 131.67 & 196.00 \\
20 & 114.00 & 102.33 & 103.67 & 169.00 \\
\hline
\end{tabular}
\vspace{1em}
\caption{Average Distance (pixels) for simulations in familiar maps}
\label{tab:avg_distance}
\begin{tabular}{|c|c|c|c|c|}
\hline
\textbf{N / d} & \textbf{0.10} & \textbf{0.20} & \textbf{0.30} & \textbf{0.40} \\
\hline
1  & 5587.74 & 5549.19 & 5512.75 & 5518.47 \\
5  & 9736.34 & 10162.66 & 10162.32 & 12148.90 \\
10 & 10621.55 & 10453.99 & 11091.13 & 16015.63 \\
15 & 12073.68 & 12486.27 & 13022.99 & 18914.38 \\
20 & 15267.22 & 13408.26 & 13843.96 & 21551.27 \\
\hline
\end{tabular}
\end{table}

\begin{figure}[!h]
\centering
\begin{minipage}{.5\textwidth}
  \centering
  \includegraphics[width=8cm]{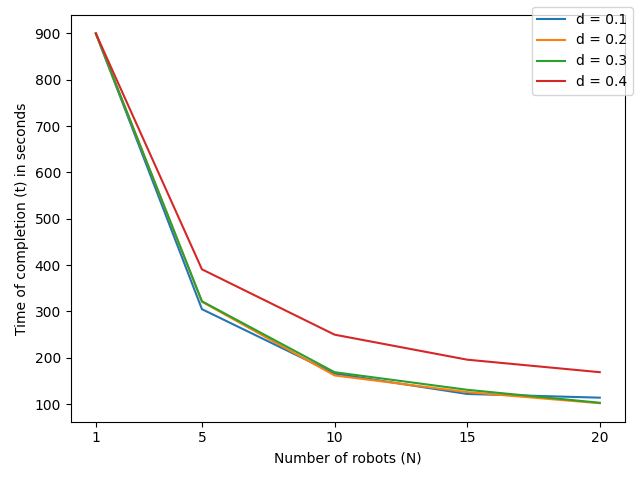}
  \caption{Completion time vs. Number of Agents.}
  \label{fig10}
\end{minipage}%
\begin{minipage}{.5\textwidth}
  \centering
  \includegraphics[width=8cm]{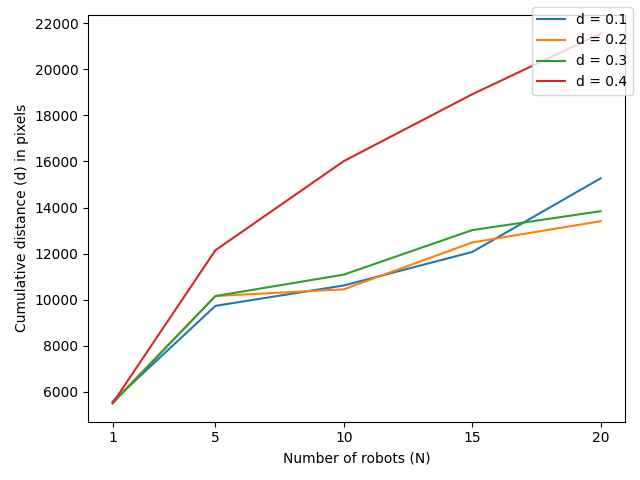}
  \caption{Cumulative distance vs. Number of Agents.}
  \label{fig11}
\end{minipage}
\end{figure}

Table~\ref{tab:avg_time} and Table~\ref{tab:avg_distance} summarize the results. As seen in Fig.~\ref{fig10}, the solution scales effectively, with completion time decreasing as agent count increases. However, the marginal gain diminishes with larger fleets due to increased collision avoidance overhead.

\begin{figure}[!h]
\centering
\includegraphics[width=12cm]{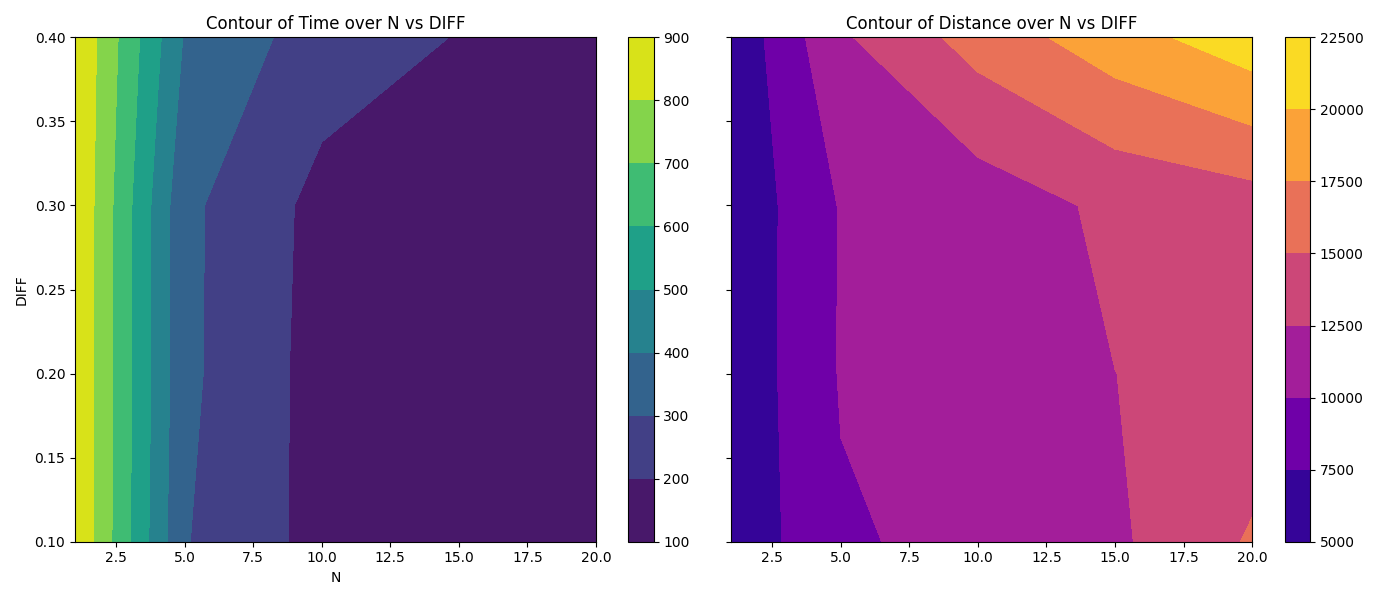}
\caption{Contour plots of Completion Time (left) and Cumulative Distance (right) over agent count vs. difficulty.}
\label{fig12}
\end{figure}

\begin{figure}[!h]
\centering
\includegraphics[width=7cm]{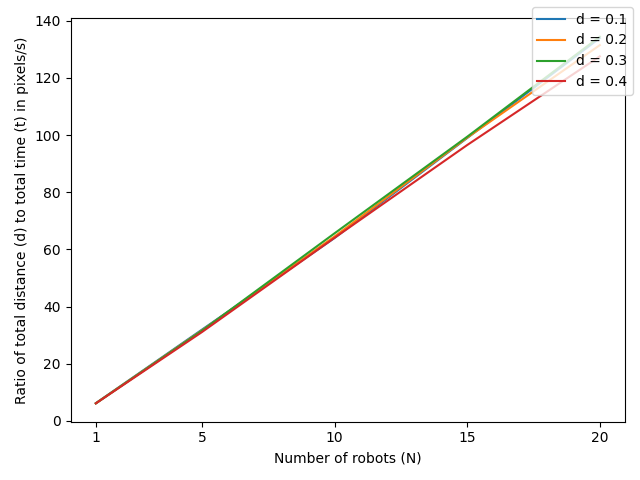}
\caption{Cumulative distance per unit completion time (pixels/s.}
\label{fig19}
\end{figure}

Fig.~\ref{fig11} and Fig.~\ref{fig12} illustrate the trade-off; more robots mean more distance traveled overall. Fig.~\ref{fig19} shows that the system scales linearly with number of robots in terms of distance penalty compared to the advantage of efficient exploration, and moreover, remains constant with respect to difficulty levels. 

\subsection{Performance on Unfamiliar Maps}

\begin{figure}[!h]
\centering
\includegraphics[width=8cm]{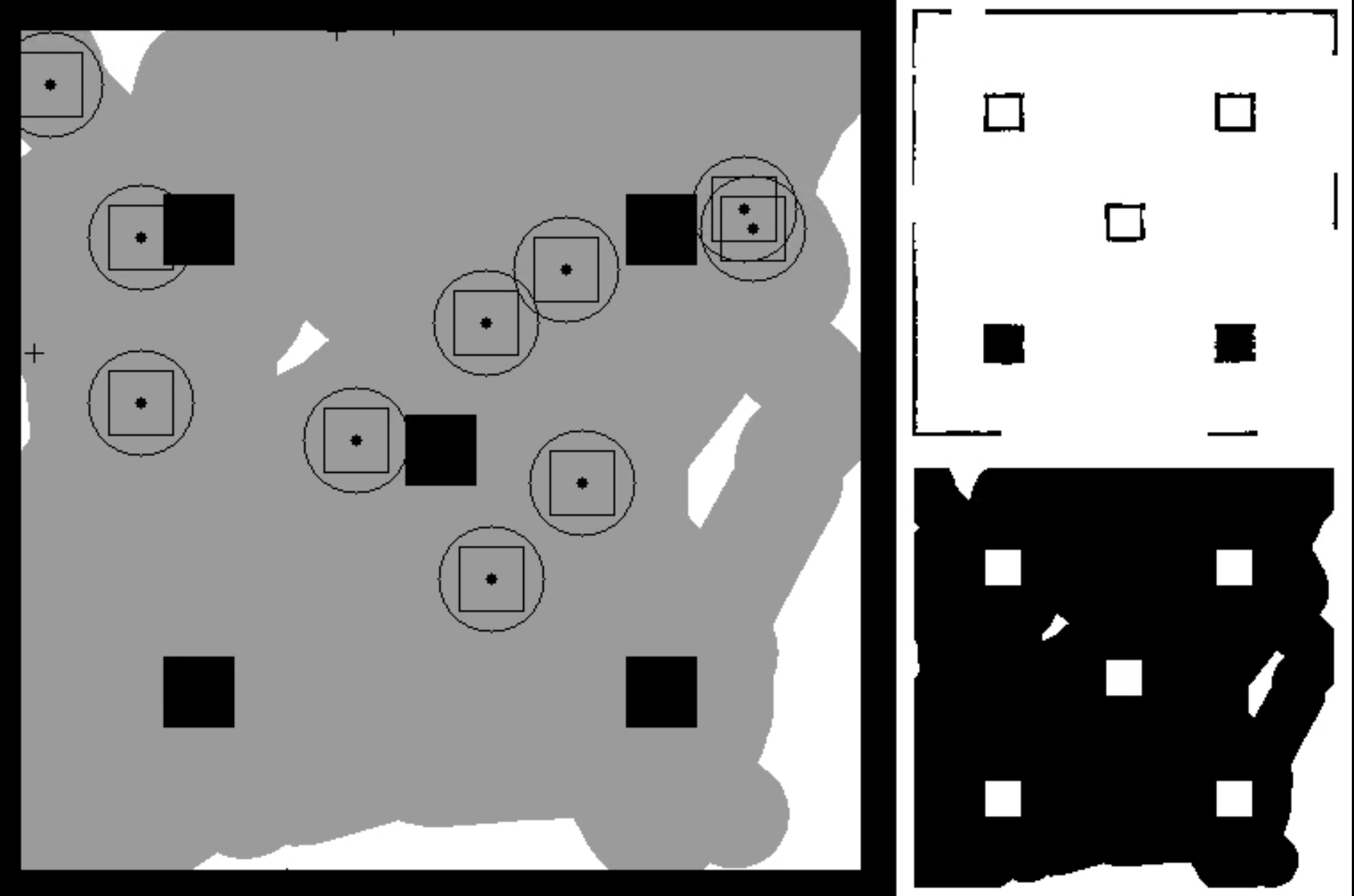}\\
\vspace{5pt}
\includegraphics[width=8cm]{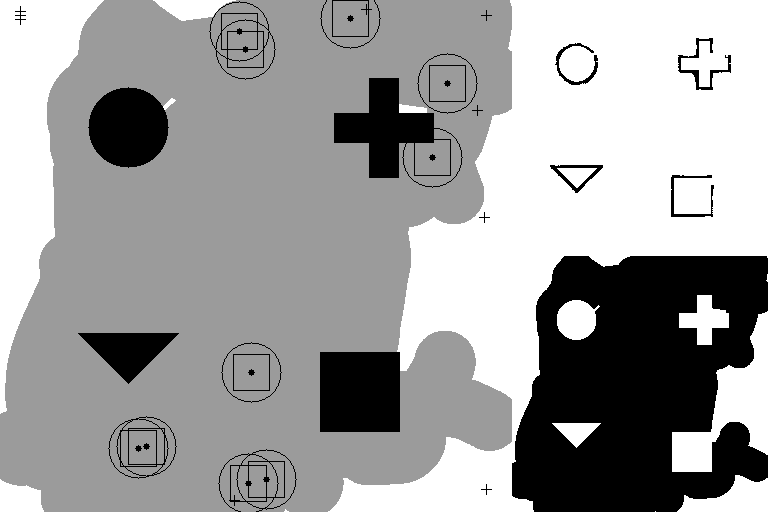}
\caption{Exploration progress in two unfamiliar maps, $N = 10$.}
\label{fig13}
\end{figure}

To test generalizability, we evaluated the system on handcrafted maps featuring geometric structures (circles, triangles, sharp corners) absent from the training data Example unfamiliar maps are visualized in Fig.~\ref{fig13}.

\begin{figure}[!h]
\centering
\includegraphics[width=\columnwidth]{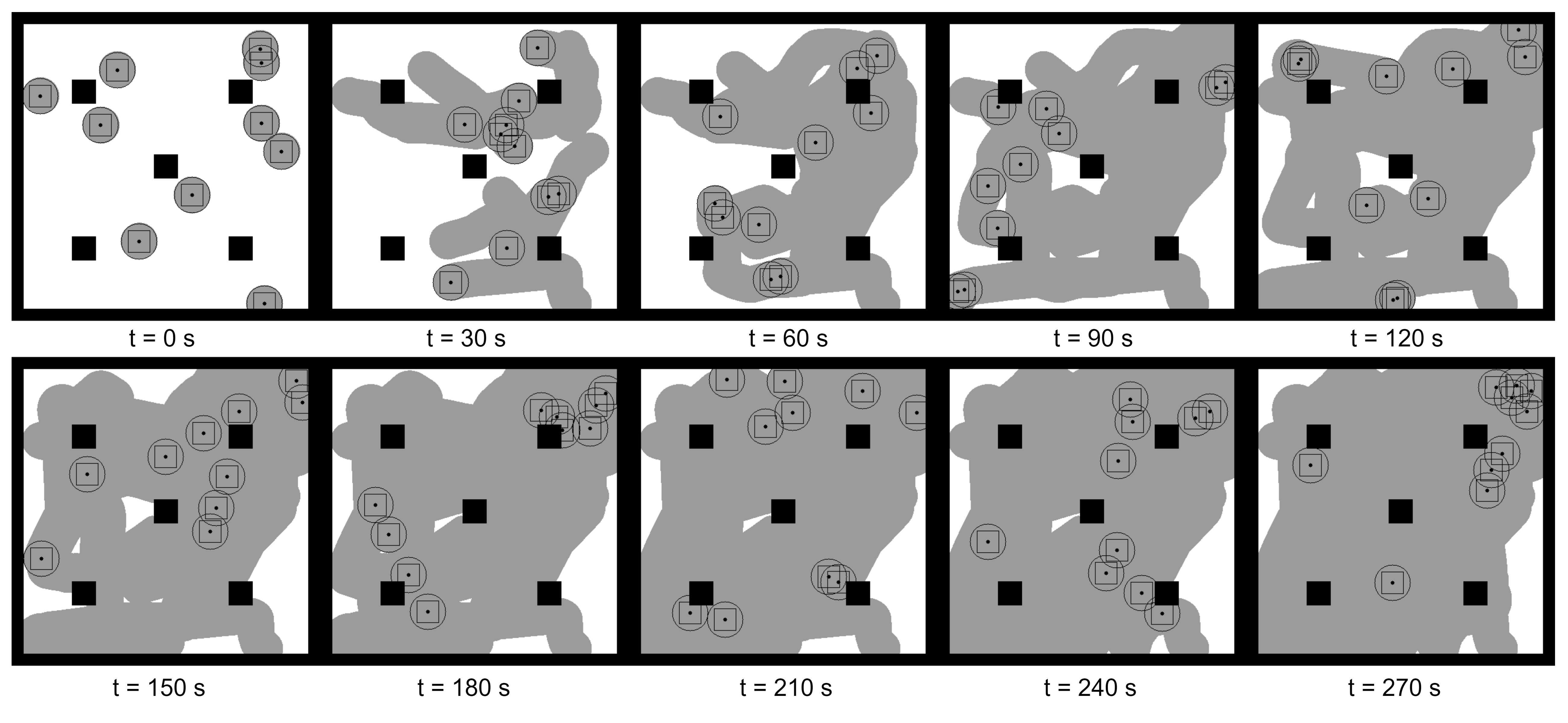}
\vspace{5pt}
\includegraphics[width=\columnwidth]{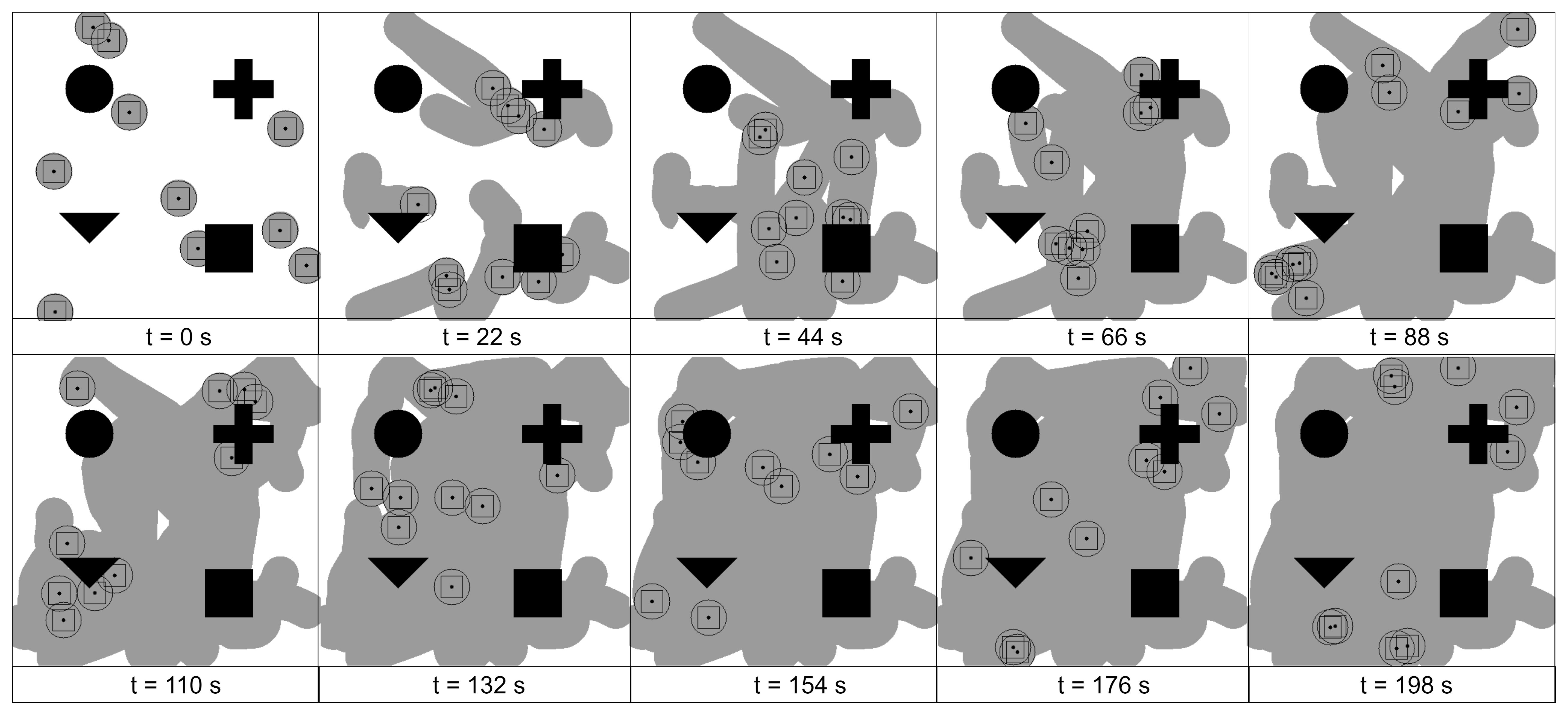}
\caption{Snapshots of exploration in unfamiliar map examples.}
\label{fig14}
\end{figure}

These maps include varying border thicknesses and regular shapes to detect overfitting. As shown in Fig.~\ref{fig14}, the policy networks and autoencoder generalize well. The performance of 10 robots on these maps closely mirrors that on familiar maps of difficulty 0.3--0.4.

\subsection{Effect of Noise}

Aside from dealing with noiseless environments where perfect communication was possible (Subsections 7.1 and 7.2), we further evaluated the resilience of the trust-based fusion mechanism by injecting Gaussian noise into inter-agent communication. The parameters are highlighted in Table 5.
\begin{equation}
\begin{split}
o_i^s &\leftarrow o_i^s + \mathcal{N}(0, \sigma^2) \\
e_i^s &\leftarrow e_i^s + \mathcal{N}(0, \sigma^2)
\end{split}
\end{equation}
where $o_i^s$ and $e_i^s$ are the encoded map vectors. Map accuracy is defined as the similarity between the ground truth and the collective obstacle map within explored regions:
\begin{equation}
\text{Accuracy} = 
\frac{
\sum_{x, y} [ ( G(x, y) = O(x, y) ) \land ( E(x, y) = 1 ) ]
}{
\sum_{x, y} [ E(x, y) = 1 ]
}
\end{equation}

\begin{table}[!h]
\centering
\caption{Parameters for Noise Tests}
\label{tab:noise_params}
\begin{tabular}{ll}
\hline
\textbf{Parameter}       & \textbf{Value}             \\
\hline
Trust Parameter ($\beta$) & 0.1, 0.3, 0.5, 0.7, 0.9 \\ 
Noise Standard Deviation ($\sigma$) & 0, 0.001, 0.002, 0.003, 0.004 \\
Number of Robots         & 10  \\
Difficulty Level         & 0.35 \\
\hline
\end{tabular}
\end{table}

\begin{figure}[!h]
\centering
\begin{minipage}{.5\textwidth}
  \centering
  \includegraphics[height=5cm]{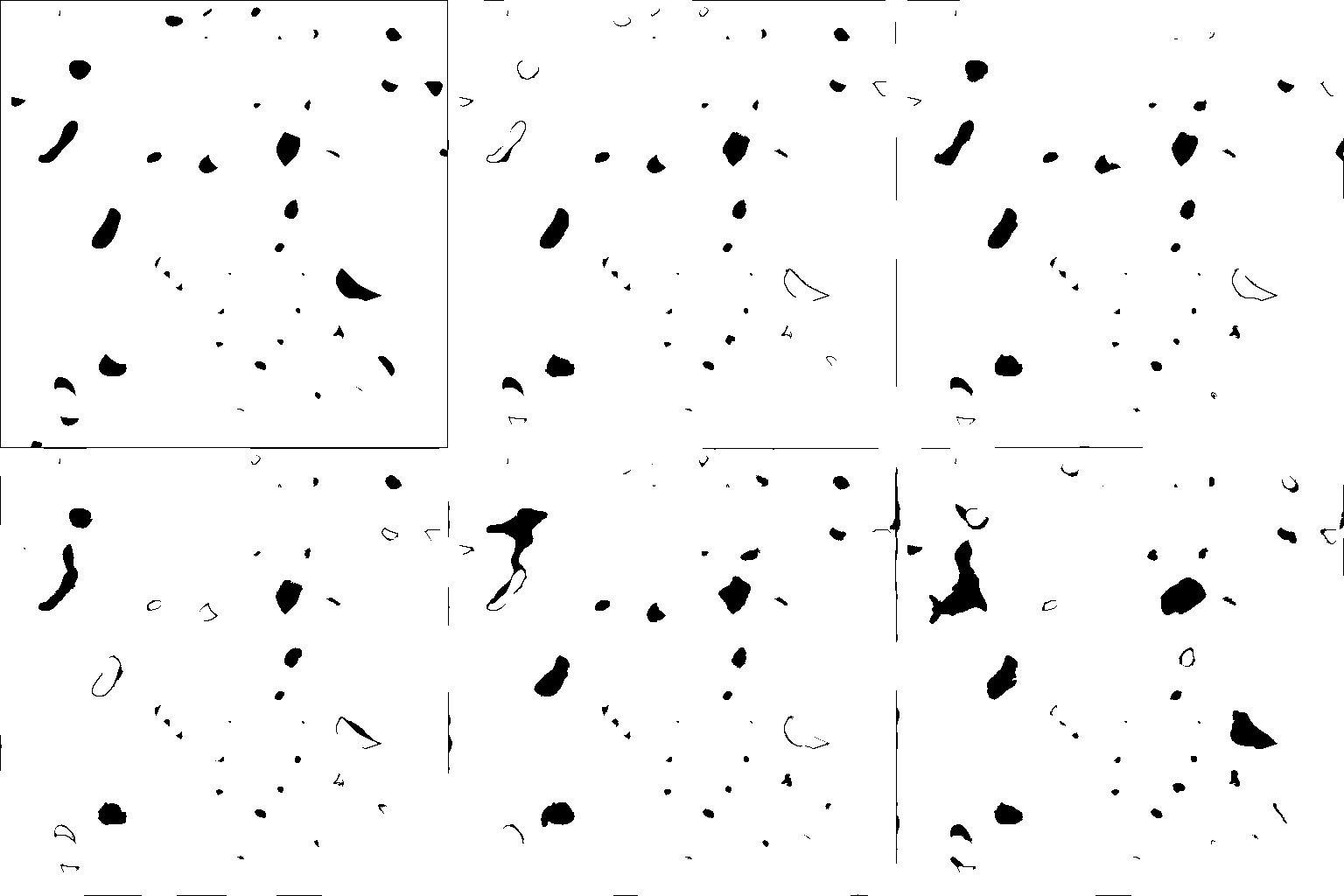}
  \caption{Effect of noise on final map quality, $\beta = 0.7$.}
  \label{fig15}
\end{minipage}%
\begin{minipage}{.5\textwidth}
  \centering
  \includegraphics[height=5cm]{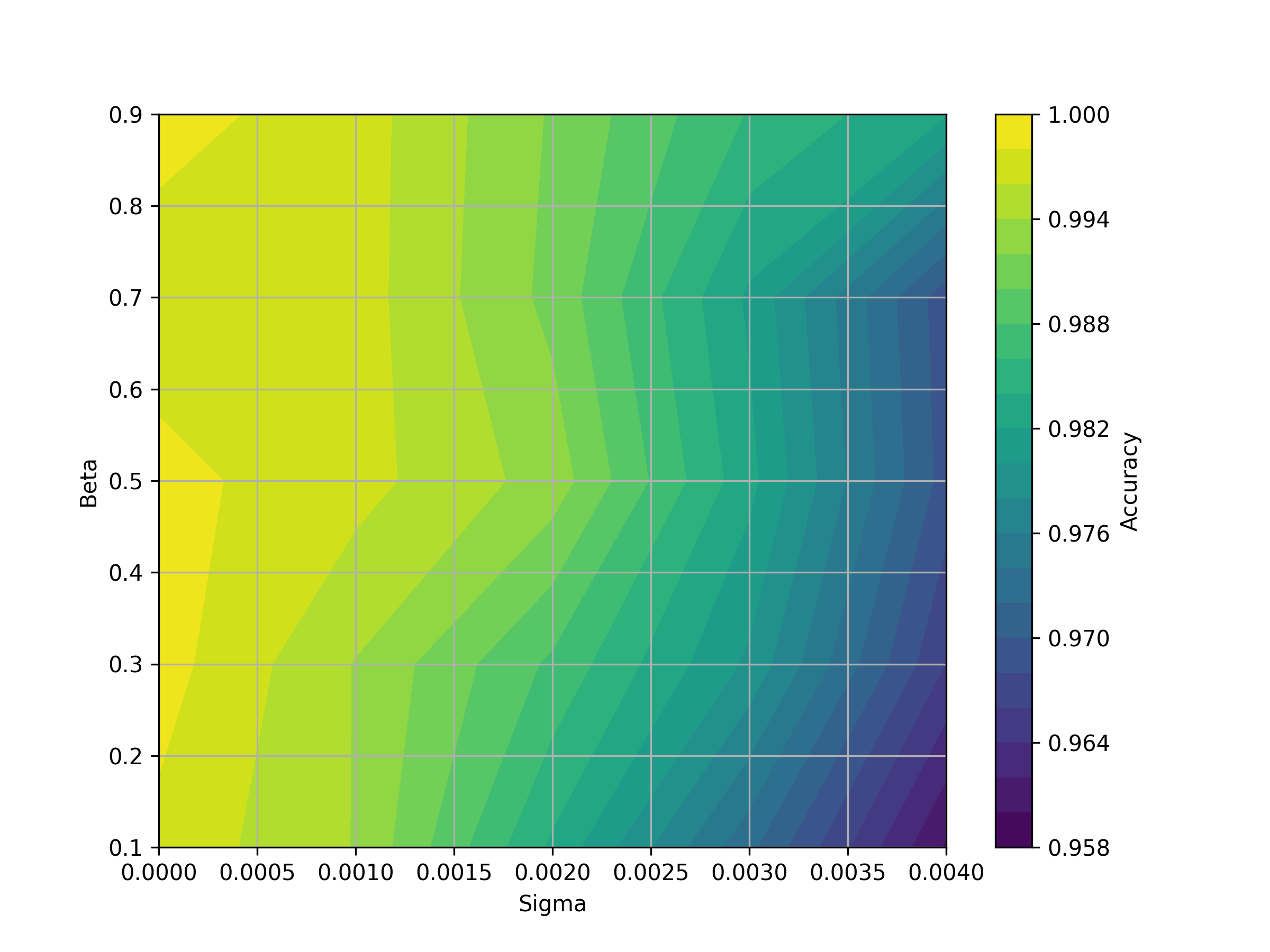}
  \caption{Contour plot of accuracy over $\beta$ vs $\sigma$.}
  \label{fig16}
\end{minipage}
\end{figure}

Fig.~\ref{fig15} visualizes the obstacle morphing caused by noise. The quantitative impact is shown in Fig.~\ref{fig16}. At low noise levels, $\beta$ has little effect. However, as noise increases, higher $\beta$ values yield significantly better accuracy. This confirms that in high-noise regimes, the fusion mechanism correctly prioritizes self-perception over noisy communication, preventing error propagation.

\section{Conclusion}
\label{sec:conclusion}
This paper presented a comprehensive framework for multi-robot exploration that addresses the critical challenges of scalability, generalizability, and noise resilience. By integrating procedural environment generation with deep reinforcement learning, we developed a system capable of navigating complex, unknown environments without relying on hand-crafted heuristics. As a major highlight, a novel layered map generation algorithm using masked Perlin noise is introduced to facilitate robust training. It produces diverse, unstructured topologies that allows for full control over obstacle size and density. The high dimensionality of mapping data is addressed through a convolutional autoencoder, which compresses occupancy grids into latent vectors, making the state space tractable for RL. Furthermore, the proposed framework employs a hierarchical architecture that allows for a decoupling between high-level planning (exploration policy) and low-level planning (navigation policy), further improving the tractability of training an RL-based solution. To ensure scalability and robustness, information sharing is restricted to self-perceived maps and a trust-based weighted consensus mechanism is introduced. It uses a tuneable trust parameter to control the influence of potentially erroneous communications. This effectively prevents the cascading of accumulation errors through the swarm, ensuring map accuracy even under noisy communication conditions.

Testing on maps generated using the same algorithm used to generate the training maps, it was observed that the system showed a steady decrease in total time taken to explore as the number of robots was increased ($1, 5, 10, 15, 20$), but accompanying this was a steady increase in the cumulative distance travelled by all the robots before concluding the exploration. This trend was observed on maps of multiple levels of difficulty ($0.1, 0.2, 0.3, 0.4$).  Importantly, it was found that the ratio of the cumulative distance travelled to the time taken only increases linearly with an increase in the number of robots. Moreover, this trend is closely followed even through increasing difficulty levels. These observations prove that the system is scalable, while tests of a similar nature performed on unfamiliar handcrafted maps with geometric structures showcased a similar pattern, proving the generalizability of the proposed solution. Therein, the implemented trust-based weighted consensus mechanism was tested for its ability to mitigate error propagation by injecting gaussian noise of varying levels into the communication links. It was found that at a noise standard deviation of $0.004$, tuning the trust parameter yielded an accuracy improvement from $95\%$ to $98\%$ on the final map.

Future work will focus on extending this framework to three-dimensional environments by utilizing 3D convolutional layers and volumetric occupancy grids. Additionally, the modular nature of the hierarchical framework allows for the adaptation of the high-level policy for specific downstream applications beyond pure exploration, such as search-and-rescue, autonomous mining, or cooperative transport operations. Finally, the individual contributions such as the Perlin noise-based map generation algorithm or the trust-based weighted consensus algorithm for information fusion can be independently used in future works tackling problems of a similar nature, such as navigation algorithms for search and rescue using swarm robots in disaster affected zones and coordination algorithms for ground coverage by cleaning or surveillance robots.

\end{document}